\documentclass[lettersize,journal]{IEEEtran}

\usepackage{amsmath,amsfonts}
\usepackage{algorithmic}
\usepackage{algorithm}
\usepackage{array}
\usepackage[caption=false,font=normalsize,labelfont=sf,textfont=sf]{subfig}
\usepackage{textcomp}
\usepackage{stfloats}
\usepackage{url}
\usepackage{verbatim}
\usepackage{graphicx}
\usepackage{xcolor}
\usepackage{cite}
\usepackage[utf8]{inputenc}
\usepackage{booktabs}
\usepackage{longtable}

\usepackage{tabularx}
\usepackage{tikz}
\usepackage{pgf-pie}
\usepackage{epsfig}
\usepackage{amssymb}
\usepackage{amsthm}
\usepackage{booktabs}
\usepackage{pgfplots}
\pgfplotsset{compat=1.18}
\usetikzlibrary{arrows.meta, positioning, shapes.geometric}
\usepackage{multirow}

\begin{document}

\title{Deep Learning for Slum Mapping in Remote \\Sensing Images: A Meta-analysis and Review}

\author {Anjali Raj, Adway Mitra\\Centre of Excellence in Artificial Intelligence, Indian Institute of Technology Kharagpur, West Bengal, India \\ Manjira Sinha\\Tata Consultancy Services Research, Kolkata, India}

\maketitle

\begin{abstract}
The major Sustainable Development Goals (SDG) 2030, set by the United Nations Development Program (UNDP), include sustainable cities and communities, no poverty, and reduced inequalities. Yet, millions of people live in slums or informal settlements under poor living conditions in many major cities around the world, especially in less developed countries. Emancipation of these settlements and their inhabitants through government intervention requires accurate data about slum location and extent. While ground survey data is the most reliable, such surveys are costly and time-consuming. An alternative is remotely sensed data obtained from very-high-resolution (VHR) imagery. With the advancement of new technology, remote sensing based mapping of slums has emerged as a prominent research area. The parallel rise of Artificial Intelligence, especially Deep Learning has added a new dimension to this field as it allows automated analysis of satellite imagery to identify complex spatial patterns associated with slums. This article offers a detailed review and meta-analysis of research on slum mapping using remote sensing imagery (2014–2024), with a special focus on deep learning approaches. Our analysis reveals a trend towards increasingly complex neural network architectures, with advancements in data preprocessing and model training techniques significantly enhancing slum identification accuracy. We have attempted to identify key methodologies that are effective across diverse geographic contexts. While acknowledging the transformative impact of Neural Networks, especially Convolutional Neural Networks (CNNs), in slum detection, our review underscores the absence of a universally optimal model, suggesting the need for context-specific adaptations. We also identify prevailing challenges in this field, such as data limitations and a lack of model explainability, and suggest potential strategies for overcoming these.
\end{abstract}


\section{Introduction}

\begin{figure}
        \centering
        \includegraphics[width=\linewidth]{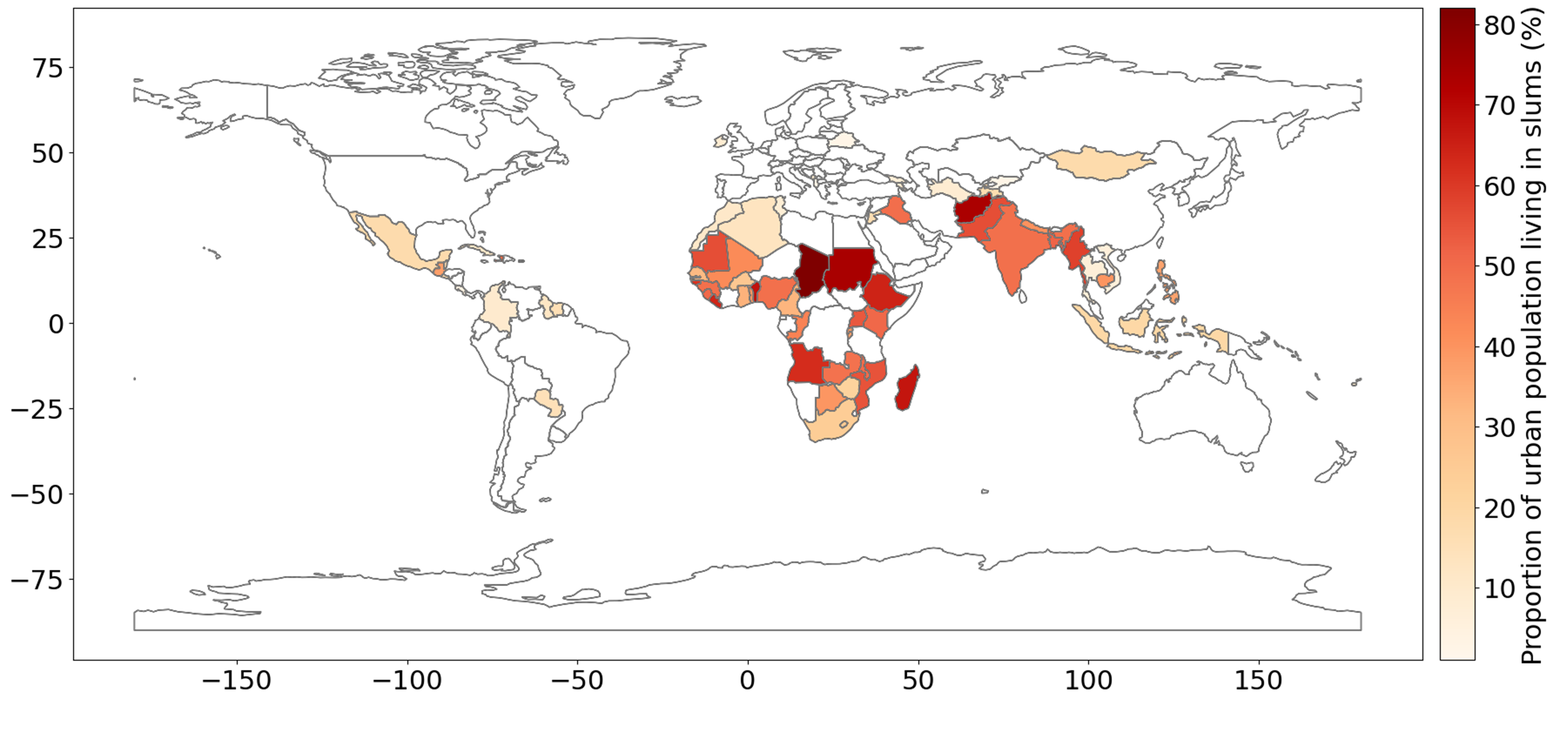}     
        \caption{Geographic distribution of the proportion of urban populations residing in slums by country \cite{unhabitatslum}. The color gradient indicates the percentage, with darker shades representing higher proportions. The absence of colors denotes the unavailability of data.}
        \label{fig:slumpop}
\end{figure}

Slums, also known as informal settlements, are areas where people live in deplorable and vulnerable conditions. They pose a significant challenge to achieving equitable and sustainable development, particularly in the context of SDG 11 \cite{sdg}, which aims to``make cities and human settlements inclusive, safe, resilient, and sustainable.'' The global population of urban slum dwellers is estimated to be one billion, disproportionately affecting the most impoverished nations \cite{UN-Habitat, UNDP}. According to recent statistical data, approximately 1.6 billion individuals out of the total urban population of four billion are residing in slum areas \cite{worldbank2021, urbanization}, and by 2050, this is expected to exceed 3 billion \cite{UNSDG2023}. These statistics highlight the urgent need to address urban poverty and improve living conditions in informal settlements. Figure ~\ref{fig:slumpop} shows the proportion of urban populations living in slums \cite{unhabitatslum}. This provides an overview of the global distribution of slums and helps identify regions where urban poverty's challenges are most severe. 

Slum dwellers face numerous social, economic, and health challenges that impact their quality of life. Socially, slums are characterized by high population density, inadequate housing, and limited access to essential services, leading to a lack of privacy and security for residents. Economically, the informal nature of slum settlements results in limited employment opportunities and financial instability. Health-wise, slums are breeding grounds for communicable diseases due to overcrowding, poor sanitation, and limited access to clean water and healthcare services \cite{unhabitat}. Climate-induced migration makes these conditions worse, highlighting the need for effective interventions \cite{climateandslums1, climateandslums2, climateandslums3}. Herein, slum detection plays a crucial role in addressing these challenges by providing accurate and up-to-date information about the location and extent of informal settlements. Precise slum mapping enables policymakers and urban planners to allocate resources effectively, prioritize infrastructure development, and deliver essential services tailored to the needs of slum dwellers. By identifying areas that require intervention, slum mapping facilitates targeted actions that can significantly improve the living conditions in these communities \cite{abascal2022domains}.  The insights gained from slum detection can integrate informal settlements into the urban fabric, promote inclusivity and resilience in cities, and contribute to the overall well-being of slum residents.

A significant obstacle to effectively managing slums and enhancing the living conditions of their inhabitants is the lack of accurate information regarding these areas. Slums evolve rapidly, while most countries conduct ground-based censuses only on a decadal basis. In light of these challenges, remote sensing emerges as a pivotal tool to shed light on the spatial characteristics and dynamics of these settlements \cite{reviewkuffer, reviewron}. The use of satellite images facilitates the mapping of slums, providing critical data for infrastructure development and urban planning. Integrating remote sensing data with different algorithms and indicators \cite{reviewkuffer, reviewron, mahabir2020ML} offers an understanding of the slum communities. These technologies allow the analysis of urban topographies, providing critical insights that support sustainable urban management and planning. For instance, models applied to satellite imagery can detect subtle changes in slum areas \cite{maiya2018slum, kit2013automated, liu2019tDL}, facilitating timely interventions to address issues of overcrowding and sanitation, thus also contributing to SDG 3 (Good Health and Well-being). 

Accurate slum mapping enables targeted interventions that can improve the living conditions in these communities, directly contributing to SDG 11. SDG 11.1 seeks to improve slums and provide secure, affordable housing and critical services to all. Accurate slum mapping identifies regions that need housing improvements and key services, enabling slum upgrading. SDG 11.3 encourages inclusive, sustainable urbanization and participatory, integrated, and sustainable human settlement planning and management. In Nairobi, Kenya \cite{mahabir2020ML}, the Kibera slum upgrading project \cite{kiberaslumupgarding} used detailed mapping to redesign the settlement, improving road networks, public facilities, and legal land tenure. This initiative enhanced resident's quality of life. It promoted participatory urban planning, and by involving the community \cite{meredith2017community} in these processes, it also supported SDG 11.3. The use of comprehensive slum maps in Maharashtra, India, \cite{sa} has improved access to sanitation and clean water, reducing disease outbreaks and also contributing to SDG 6 (Clean Water and Sanitation). This effort has improved housing developments, making them safer and more durable, and has also supported the inclusion of slum dwellers in urban planning discussions. This shift towards data-driven, participatory planning has improved living conditions and also fostered a sense of community ownership and empowerment among residents, crucial for sustainable urban development. 

The IDEAMAPS Data Ecosystem \cite{ideamaps} enhances slum mapping through artificial intelligence and community engagement, supporting informed urban planning and contributing to SDG 11. The SLUMAP project \cite{slummaps} uses remote sensing to map slums in sub-Saharan Africa, influencing policies for SDG 11 and SDG 3. Such technologies allow precise interventions to improve living conditions and promote inclusive urban growth, making cities safer and sustainable. These efforts highlight the importance of community involvement and data-driven planning in transforming slums into integrated, resilient urban spaces. Hence, the advancements in slum mapping technologies are not just academic; they translate into tangible benefits that enhance the quality of life of millions living in marginalized urban areas. Thus, slum mapping is critical to achieving SDG 11 by providing the foundation for informed and inclusive urban planning and policy-making. 

\begin{figure*}[!h]
        \centering
        \includegraphics[width=\linewidth]{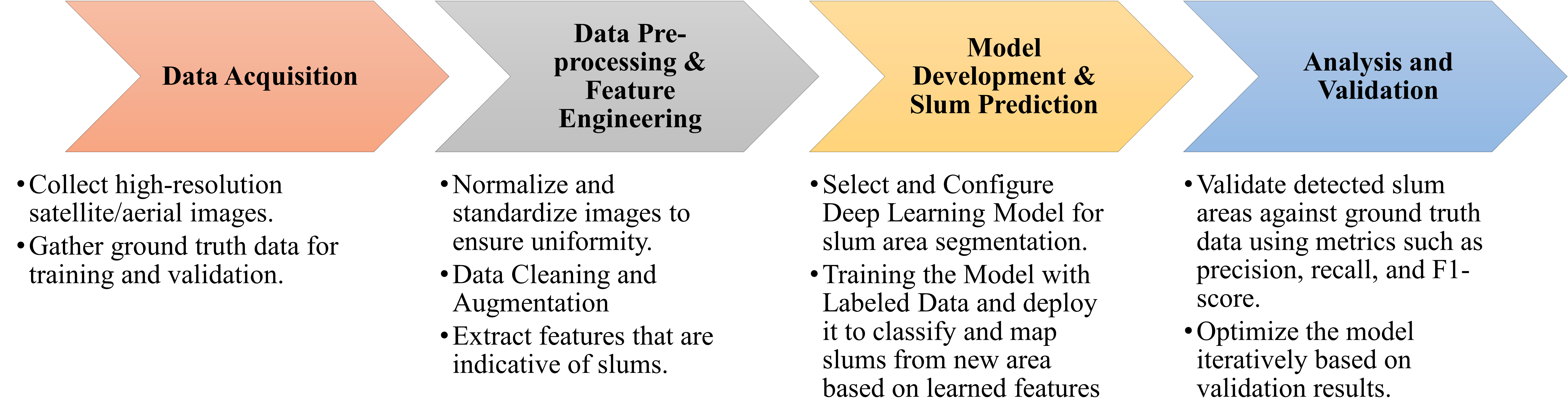} 
        \caption{Comprehensive Workflow of Deep Learning for Slum Mapping. This diagram highlights the essential stages in the deep learning process used for effective slum detection and analysis}
        \label{fig:workflow}
\end{figure*}

As technology evolves, machine learning and deep learning have become crucial to harness remotely sensed data \cite{zhu2017deep, yuan2021review}. These algorithms can process complex, high-dimensional imagery and extract intricate patterns often invisible to human analysts, enhancing the precision and granularity of slum detection. This study explores the relationship between remote sensing and deep learning within the specific domain of slum mapping, critically evaluating the deep learning architectures employed in this context. A structured approach leveraging deep learning techniques is essential to effectively tackle the challenges associated with mapping slums. Figure ~\ref{fig:workflow} illustrates the workflow adopted in this study. This workflow underscores the methodical steps involved in deploying deep learning techniques for slum mapping, setting the foundation for the detailed exploration of methodologies discussed in the following sections.

The subsequent sections of this paper are structured as follows: Section II provides an overview of several methodologies employed in slum mapping through remote sensing imagery. Section III offers a concise overview of deep learning architectures. Section IV presents a comprehensive meta-analysis of deep learning-based algorithms used for slum mapping and identifies a few challenges to focus on. Section V examines the issues mentioned in the preceding section in detail and discusses the approaches taken in the literature. Section VI explores integrating deep learning with Geographic Information Systems (GIS) software. We discuss various ethical concerns and limitations of remote sensing in Section VII and draw our conclusions in Section VIII.

\section{Slum Mapping using Remote Sensing}

\subsection{Technological Evolution in Slum Mapping: From Ground Surveys to Remote Sensing}

Slum mapping has evolved from labor-intensive census surveys \cite{baud2009matching, weeks2007can} and data gathering to the advanced analysis of satellite imagery. Modern remote sensing technologies, like the Sentinel, Ikonos, and WorldView satellites, have changed this field by providing moderate to very high-resolution (VHR) imagery (In this study, we define VHR images as those with a spatial resolution of 1 m or less, high-resolution images with a spatial resolution between 1 m and 5 m, and moderate-resolution images with a spatial resolution between 5 m and 10m. A spatial resolution of over 10 m is regarded as a low-resolution image). This has greatly improved urban data processing \cite{kit2013automated, jacobsen2008mapping, kuffer2016texture}, made images more detailed, and helped researchers figure out the complex urban features of slum areas.

In addition to satellite imagery, aerial imaging techniques have seen substantial progress. Unmanned Aerial Vehicles (UAVs), or drones, have become increasingly popular for capturing high-resolution images of urban environments \cite{gevaert2017ML}. Furthermore, the integration of advanced sensors on drones and aircraft provides high-resolution data crucial for delineating intricate slum features. The integration of LiDAR technology with remote sensing has further enhanced the accuracy of slum mapping \cite{ribeiro2019object}. LiDAR sensors, mounted on satellites, aircraft, or drones, use laser pulses to create detailed representations of urban environments and capture the topography of slum areas.

Before adopting machine learning approaches, object-based image analysis (OBIA) and texture analysis were commonly employed to identify slum areas. These methods, discussed in detail in the subsequent subsections, have laid the groundwork for the current state-of-the-art approach to slum mapping. The transition from conducting surveys on-site to using remote sensing techniques represents a shift towards an efficient, scalable, and objective approach to analyzing urban landscapes \cite{moeller, fugate2010survey}. This technological evolution has paved the way for creating specific policies and interventions that cater to the ever-evolving needs of metropolitan regions, contributing to effective urban planning and policy-making. 

\subsection{Texture Analysis in Slum Mapping}

Texture analysis plays a vital role in remote sensing for slum mapping, as it helps to distinguish informal settlements from formal urban areas based on their unique spatial configurations and architectural patterns. In this section, we emphasize the importance of texture analysis in slum mapping and present studies where this approach has achieved successful results.

High-resolution satellite images reveal textural contrasts between slums and formal urban areas, serving as a key identifier for remote sensing techniques. There are two primary approaches for texture analysis: structural methods like mathematical morphology (MM) \cite{ioannidis2009towards, aptoula2011morphological}, and statistical methods like the Grey Level Co-occurrence Matrix (GLCM) \cite{glcm}. Structural approaches analyze images by examining geometric structures, while statistical techniques focus on pixel-intensity arrangements to identify disordered texture patterns.

Several studies have demonstrated the effectiveness of integrating spectral information with texture analysis to improve classification accuracy. For instance, \cite{kuffer2016texture} integrated spectral information with GLCM variance to differentiate slums from formal regions. \cite{shabat2017texture} found that local directional pattern (LDP) techniques outperformed GLCM in categorizing informal settlements due to LDP's sensitivity to texture directional variance. \cite{wurm2017texture} employed a Random Forest (RF) classifier with GLCM and differential morphological profiles (DMP) to detect slums, highlighting the power of machine learning algorithms in analyzing complex texture-based data for precise slum mapping. \cite{prabhu2018texture} compared statistical and spectral features and found that spectral characteristics better described urban slum textures than statistical descriptors. \cite{wang2019texture} found that scale and anisotropy affect texture indices, emphasizing the need to select appropriate scales for slum textures. \cite{prabhu2021texture} expanded upon previous research by incorporating statistical, spectral, and structural techniques, including morphological profiles (MP) and mathematical morphological profiles (MMP), to differentiate different textures present in slum areas. They employed the minimum redundancy maximum relevance (mRMR) feature selection method before using a support vector machine (SVM) classifier. These studies demonstrate the importance of texture analysis in slum mapping, as it enables the extraction of meaningful information from high-resolution satellite images to distinguish slums from formal urban areas accurately. The methodologies used in these studies, including GLCM, LDP, and morphological profiles, highlight the diversity of approaches available for texture analysis in remote sensing applications.

While texture analysis has improved slum detection, challenges persist. The texture measures for slums can differ between locations, even within the same slum. This can be explained by variations in slum characteristics, such as the physical dimensions, layout, and materials used in construction \cite{graesser2012image, gholoobi2015using}. Thus, patterns in one image may not work for another image as image texture does not analyze pixels individually. Instead, it focuses on groups of pixels that create distinct patterns to distinguish features. However, satellite imaging and machine intelligence offer exceptional opportunities to enhance urban planning and resource allocation. Researchers and practitioners can attain detailed and flexible comprehension of informal urban settlements by combining deep learning techniques with texture analysis. Deep learning models' automatic feature learning and multiscale analysis can improve the precision and effectiveness of slum mapping, leading to targeted and effective interventions for urban development.

\subsection{Object-Based Image Analysis (OBIA)}

OBIA considers an image to be a collection of objects and uses attributes such as size, shape, texture, and relationships with neighboring objects to extract meaningful information. OBIA, in slum mapping, segments satellite images into meaningful objects or groups of pixels. These objects are then analyzed based on their spatial, spectral, and textural properties. 

One of the earliest studies employing OBIA on slums was by \cite{hofmann2001object}. This study used eCognition software \cite{baatz2001ecognition} to differentiate informal settlements from other land-use types based on their unique attributes in Ikonos images over Cape Town. The eCognition software adheres to the principles of object-oriented programming. Additionally, it has a patented methodology for image segmentation that operates across many scales. However, the approach was complex and tailored to a specific dataset, posing challenges in generalizability. Subsequent studies have built upon \cite{ hofmann2001object}, employing ontology-based approaches and refining the classification of slums \cite{ hofmann2008object, niebergall2007object,veljanovski2012object}. \cite{kohli2012ontology} presented a thorough slum ontology known as the generic slum ontology (GSO). This ontology organizes concepts into three levels: environment, settlement, and object, providing a framework for classifying slums based on image data. This ontology-based approach has been applied in other studies \cite{kohli2016texture}, based on the vegetation, impervious surface, and bare soil (V-I-S) model \cite{ridd1995VISmodel}. The advancements in OBIA have seen the integration of additional indicators and image-based features to improve the accuracy of slum mapping\cite{pratomo2017object}. \cite{mudau2021object} incorporated indicators such as plant cover, built-up area, iron cover, asphalt cover, and texture, along with NDVI, BAI, iron index, REI, coastal blue index, and texture for classification. \cite{georganos2021object} used WorldView-3 imagery and a Geographic Object-Based Image Analysis (GEOBIA) processing chain framework to classify land cover in impoverished urban areas of Nairobi, Kenya, demonstrating the effectiveness of OBIA in capturing land cover characteristics in slum contexts. These studies demonstrate the diverse applications and effectiveness of OBIA in accurately identifying and characterizing informal settlements. 

The progress in OBIA offers significant prospects for improving the accuracy and granularity of slum mapping. Nevertheless, there are still obstacles when applying findings to a wide range of situations and effectively handling intricate information.  An issue arises when vegetation and shadows obstruct some parts or entire buildings, resulting in decreased accuracy \cite{galeon2008estimation, novack2010urban}. Another concern with OBIA is the high level of spectral noise caused by the materials used in the construction of slum houses. Unpaved roads have spectral reflectance similar to slum rooftops, making it difficult to differentiate dwellings. Furthermore, the algorithms used to identify slums in an image are context-specific, restricting their applicability in other geographical regions \cite{owen2013approach}. By harnessing the capabilities of deep learning, OBIA can provide more precise and comprehensive identification of slum areas, improving urban planning and policymaking.

\subsection{Data Mining Techniques for Slum Identification}

Data mining techniques have become increasingly popular in recent years for identifying slums. These methodologies use various tools to uncover new patterns in large datasets, with machine learning and deep learning being the cornerstone of these strategies.

Several studies have applied data mining techniques to slum mapping. \cite{kemper2015ML} developed an approach using symbolic machine learning and association analysis to extract image data from SPOT-5 satellite imagery in South Africa. \cite{duque2017ML} employed logistic regression (LR), SVM, and RF algorithms to classify urban areas as slums or non-slums using spectral, texture, and structural properties extracted from Google Earth imagery. \cite{kuffer2017ML} used WorldView-2 images and auxiliary spatial data to identify deprived regions in Mumbai, employing both RF classifiers and LR models. \cite{leonita2018ML} tested SVM and RF for slum mapping in Bandung, Indonesia, using features extracted from a local slum ontology. \cite{ranguelova2019ML} used the Bag of Visual Words framework and Speeded-Up Robust Features (SURF) for pixel-level classification of slums in Kalyan and Bangalore, India.\cite{mahabir2020ML} employed discriminant analysis, LR, and See5 decision trees to assess the spatial extent and changes of slums in three Kenyan sites.

Deep learning techniques have also been explored for slum mapping. \cite{mboga2017DL} used CNNs to identify informal settlements, comparing the CNN model against SVM using GLCM and LBP features. \cite{wang2019deprivation} employed deep convolutional networks for detecting small-scale deprivation zones in Bangalore using a U-Net-Compound model for accurate city poverty mapping. \cite{ajami2019identifying} examined the relationship between deprivation and image-based traits in different slums using CNN-learned spatial features. In \cite{wurm2019}, transfer learning was used to train a model on QuickBird images and apply it to Sentinel-2 and TerraSAR-X data for slum mapping. Transfer learning has also been used in \cite{stark2020, verma2019}. \cite{liu2019tDL} suggests using fully convolutional networks (FCNs) with dilated convolutions to analyze temporal trends in transient slum clusters in Bangalore. In \cite{prabhu2021DLtexture}, a dilated kernel-based deep CNN (DK-DCNN) technique was used to detect urban slums in Indian cities, using GLCM texture features and WFT-based spectral features for classification.

Data mining has been used to identify informal settlements using UAV features \cite{gevaert2017ML}, SAR data \cite{schmitt2018texture, wurm2017textureSAR}, and Lidar data \cite{ribeiro2019object}, showcasing the versatility of slum mapping in diverse urban landscapes. To improve urban settlement mapping, \cite{lai2020texture} categorized land cover into six types using Landsat-8 and non-spectral data (digital elevation models and road networks). \cite{friesen2018size} examined the rank-size distributions of morphological slums in different cities to determine if they are comparable. They found that typical patterns found between cities can be applied to patterns within cities. \cite{friesen2019object} found morphological slums in eight cities in Africa, South America, and Asia and looked at how their sizes were spread based on \cite{friesen2018size}. \cite{kohli2016ML} examined slum identification and delineation deviations in VHR images from Ahmedabad (India), Nairobi (Kenya), and Cape Town (South Africa). Existential and extensional uncertainty in slums were shown with random sets, and bootstrapping was used to find confidence in the different definitions. They also found the built-environment criteria that experts use to spot slums in areas that do not normally have slums. The IMMerSe (integrated methodology for mapping and classifying precarious settlements) methodology was used to identify precarious settlements in the Baixada Santista metropolitan region of So Paulo \cite{da2021ML}. It characterized the urban environment using high-spatial-resolution images without digital image processing by extracting density and urban organization level. \cite{kuffer2021ML} proposed the systematic semi-automated SLUMAP framework based on free, open-source software that gives policymakers information about poor urban areas in Sub-Saharan Africa. The application of RF on Landsat 8 has been tested in \cite{assarkhaniki2021ML} using two approaches for developing the training set: using available databases (maps for informal settlements) and visual interpretation using VHR. OpenStreetMap (OSM) was used in the secondary approach to build the sample set for the classes with the lowest accuracy and precision in the first round of classification. According to \cite{ibrahim2021DL}, a realistic-dynamic urban modeling system can detect informality, slums, and pedestrian and transit modes in urban landscapes using aerial and street view images. The model uses two deep CNNs pre-trained differently on various data sets to extract and geo-reference information from unlabeled urban scene images from around the world.  The use of different data sources, from satellite imagery to Lidar data, highlights the adaptability of these methodologies. 

Data mining techniques are increasingly integrated with other technologies, such as OBIA and GIS, to enhance slum mapping \cite{dos2022identifying, priyadarshini2020identification, farooq2022slum}. This integration uses each approach's strengths to analyze informal settlements more thoroughly and accurately. For instance, using OBIA and machine learning enables the classification of complex urban structures by segmenting high-resolution images into meaningful objects and then applying data mining algorithms to these objects. \cite{fallatah2020ML} employed this approach to define slums in Jeddah. This integration allows for a more detailed understanding of slum areas, facilitating the development of targeted interventions. Similarly, GIS technologies are combined with data mining techniques to incorporate spatial analysis into slum mapping. GIS provides a framework for managing and analyzing spatial data, while data mining techniques can uncover patterns and relationships within this data. By integrating these two approaches, researchers can create more sophisticated models that consider the spatial distribution of slums and their relationship to other urban features.

Machine learning and deep learning techniques have become increasingly popular in slum mapping due to their ability to handle complex data and extract meaningful patterns. These techniques can be applied in various ways.
\begin{itemize}
    \item Feature Extraction: Machine learning algorithms can automatically identify relevant features from satellite imagery that distinguish slums from other urban areas. Features such as texture, shape, and spectral signatures are commonly used.
    \item Classification: Once features are extracted, machine learning classifiers, such as RF, SVMs, and LR, can classify areas as slums or non-slums.
    \item Deep Learning Architectures: Deep learning models, particularly CNNs, can automatically learn hierarchical feature representations from raw imagery data. This is useful for capturing the complex spatial patterns of slums.
    \item Transfer Learning: In situations where labeled data is scarce, transfer learning can be applied to adapt pre-trained deep learning models to slum mapping, leveraging knowledge from other domains.
\end{itemize}

The implications of these findings for future research and policy-making are significant. Accurate slum maps generated through these advanced techniques can inform targeted interventions for slum upgrading, infrastructure development, and service provision, contributing to the sustainable development of urban areas.

\subsection{Recommendations and Comparative Analysis for Slum Mapping Approaches}

Based on various remote sensing techniques for slum mapping, we provide the following practical guidance and comparative analysis to assist researchers and practitioners in selecting the most suitable approach for their projects:
\begin{itemize}
    \item Utilize advanced satellite systems like WorldView for large-scale projects requiring very high spatial resolution. For more detailed imagery, consider aerial imaging techniques with drones.
    \item Employ texture analysis methods, such as GLCM or mathematical morphology, to capture the distinct architectural patterns of slums, especially in complex urban environments.
    \item Use OBIA for detailed feature analysis and to reduce classification noise. This approach is particularly useful for projects that require a nuanced understanding of slum features.
    \item Automate feature extraction and classification with machine learning and deep learning techniques. These approaches help handle large datasets and capture complex spatial patterns.
\end{itemize}

\section{Unveiling Architectures: A Dive into Convolutional Neural Networks and Autoencoders}

Deep learning \cite{goodfellow2016deep} has been recognized as one of the ten groundbreaking technologies of 2013 \cite{dl}. It involves neural networks (NN) with multiple hidden layers, distinguishing it from shallower learning models. The training procedure for the deep learning model involves forward propagation, where the model predicts outputs from inputs, and backpropagation, where errors are minimized by adjusting weights and biases. Top companies like Google, Microsoft, and Facebook have invested in deep learning for applications such as image segmentation and object detection, demonstrating superior performance in complex computational tasks. Due to its remarkable achievements, it has become increasingly popular as the preferred model in numerous application domains. Following this achievement and the enhanced accessibility of data and computational resources, it is gaining momentum even in remote sensing.

Deep learning has enhanced the capabilities of remote sensing and urban analysis through the processing and analysis of complex data. Remote sensing uses deep learning algorithms for tasks such as land cover classification \cite{kussul2017deep, helber2019eurosat, vali2020deep}, object detection \cite{li2020object, zhao2019object, mohamed2022building}, and change detection \cite{wen2021change, wang2021mask, ma2023dual}. These techniques excel at evaluating high-resolution satellite and aerial images, allowing for precise and detailed urban mapping. Within urban analysis, it allows for the automated extraction of complex patterns present in densely populated urban areas, helping in detecting slums. This section provides a concise overview of deep-learning architectures employed in slum mapping.

\subsection{Convolutional neural networks (CNNs)}

\begin{figure}[h]
        \centering
        \includegraphics[width=\linewidth]{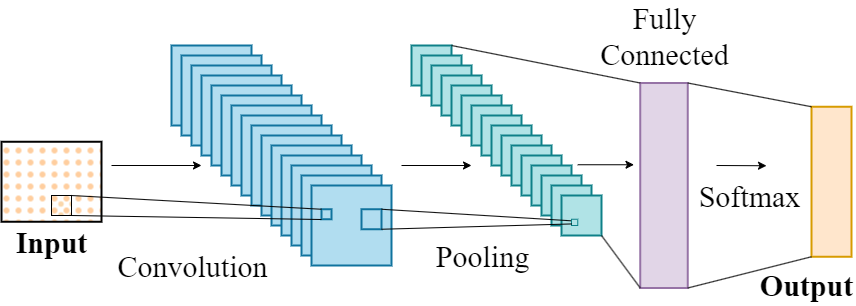}       
        \caption{Architecture of a Convolutional Neural Network (CNN)}
        \label{fig:cnn}
\end{figure}

Yann LeCun introduced convolutional networks in 1989 \cite{lecun}, which fundamentally consist of three stages: convolution to generate linear activations, application of nonlinear activation functions like ReLU \cite{relu}, and pooling \cite{pooling} to reduce dimensionality. These layers, along with fully connected layers, help CNNs capture complex hierarchical image representations. Figure ~\ref{fig:cnn} shows the basic architecture of a CNN. Notable for their versatility and scalability, CNNs have excelled in various computer vision tasks. For instance, AlexNet \cite{alexnet} won the 2012 ImageNet challenge, demonstrating the power of deep CNNs. VGG networks \cite{vgg} improved classification by deepening network structures, while ResNet \cite{resnet} introduced skip connections to ease training deep networks. Fully Convolutional Networks (FCNs) \cite{fcnn} use an encoder-decoder structure to handle various input sizes, crucial for tasks like semantic segmentation.

\subsection{Autoencoders}

\begin{figure}[h]
        \centering
        \includegraphics[width=\linewidth]{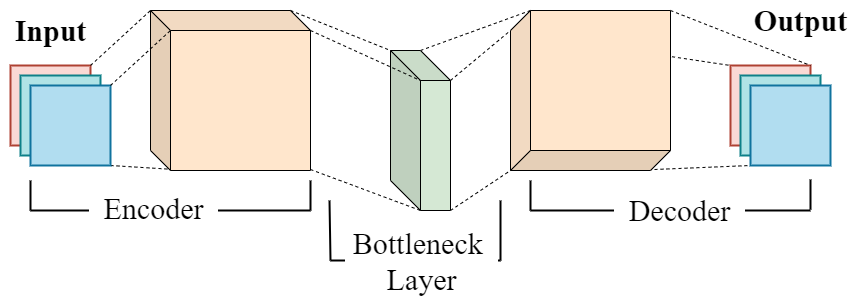}        
        \caption{Encoder-Decoder structure of an Autoencoder.}
        \label{fig:autoencoder}
\end{figure}

An autoencoder is a NN designed to replicate its input data at its output. It consists of an encoder function $h = f (x)$ that compresses the input into a lower-dimensional latent space and a decoder function $r = g (h)$ that reconstructs the input from this compressed representation. Figure ~\ref{fig:autoencoder} shows the basic encoder-decoder structure of an autoencoder. They have been particularly effective in remote sensing for feature representation, as noted in \cite{aers1, aers2, aers3}.

\section{Meta-Analysis}

The use of meta-analysis has emerged as a fundamental methodology in numerous academic fields. It allows researchers to combine data and conclusions from several studies to gain broader insights that may not be readily evident from any single study. The word ``meta-analysis'' was first used in academia to organize and consolidate the examination and amalgamation of numerous studies. \cite{glass1976} defines it as a systematic procedure for reviewing and combining various analyses. Over time, meta-analysis has undergone a process of evolution and refinement. In essence, it is not simply a compilation of several studies but rather a rigorous methodology that integrates findings from comparable studies, as detailed by \cite{eggermetaanalysis}. The selection and evaluation of pertinent studies are fundamental components in the execution of a meta-analysis. Researchers often use established procedures to select studies uniformly, clearly, and thoroughly. One example of a generally acknowledged standard is the Preferred Reporting Items for Systematic Reviews and Meta-Analyses (PRISMA) \cite{prisma2009}. PRISMA, a framework developed to assist researchers in conducting systematic reviews \cite{king2005} and meta-analyses, places significant emphasis on the importance of clear and comprehensive reporting. The use of the PRISMA protocol is not limited to any discipline. For example, research conducted by \cite{khatami2016} and \cite{george2019} has demonstrated the PRISMA protocol's applicability in urban geography. Different perspectives can be gained by using meta-analysis in scholarly research. This helps researchers find recurring patterns, come up with stronger conclusions, and propose recommendations based on a wider range of data. In the context of slum mapping using deep learning, meta-analysis is particularly relevant due to the diverse methodologies, varying data sources, and the need for consolidated insights to guide future research and applications.

\subsection{Selection of scientific articles} 

\begin{figure}[htb]
        \centering
        \includegraphics[width=\linewidth]{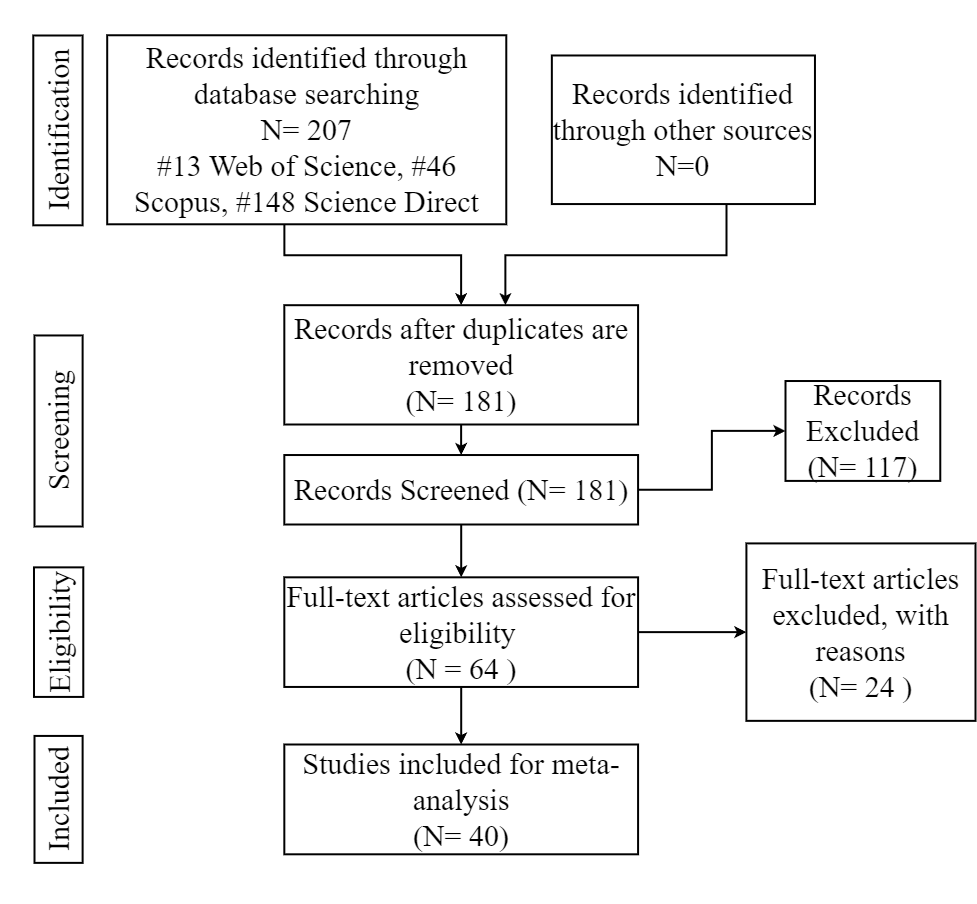}        
        \caption{PRISMA flow diagram depicting the systematic selection process of the articles.}
        \label{fig:datasearch}
\end{figure}

In this information age, the methodology employed to sift through the vast literature is of the utmost significance. The objective of this study is to compile and evaluate studies that use deep learning in the context of mapping slums. This section provides an account of the approach employed for selecting and reviewing the papers. The study used scholarly articles obtained from three widely recognized academic databases, namely Web of Science, Scopus, and Science Direct. These databases are well-known for their extensive compilation of high-quality research articles spanning several academic fields. The publications chosen for this study were limited during the timeframe of 2014–2024. This selection was made to provide a thorough understanding of the latest developments in the field. A precise search query was created to ensure the articles were pertinent to the study's subject. The search focused on the database record title, abstract, and keywords sections. The search query performed consisted of the terms: ``slum'' OR ``informal settlement'' AND ``remote sensing'' OR ``satellite imagery'' AND ``deep learning'' OR ``neural networks''. The search was done on April 5, 2024. After performing the search query, the database yielded a cumulative count of 207 publications. However, not all these papers were relevant for our review. Consequently, further filtering procedures were implemented; review articles were excluded. The scope of this study was limited to papers authored exclusively in English. An examination of article duplication was conducted, and redundant articles were removed. Exclusions were made for grey literature, book sections or chapters, reports, and thesis publications. Also, studies that did not pertain directly to using deep learning techniques in examining slums were excluded. Studies without sufficient methodological details or validation processes were also excluded. Only studies that provided full-text access were included. Any articles inaccessible due to a paywall or lack of institutional access were removed from the analysis. Studies on change detection, land use, land cover, health, sanitation facilities, and settlement expansion were omitted. After filtering, 40 publications \cite{wahbi2023deep, lu2024geoscience, chen2022hierarchical, abascal2024ai, el2023building, lumban2023comparison, huang2023comprehensive, persello2017deep, wang2019deprivation, stark2023detecting, debray2019detection, mboga2017DL, wang2023eo+, dabra2023, fan2022fine, bardhan2022geoinfuse, ajami2019identifying, abascal2022identifying, ansari2020, najmi2022integrating, gadiraju2018machine, raj2023mapping, mattos2022mapping, rehman2022mapping, fan2022multilevel, el2024new, stark2020, wurm2019, el2023slum, stark2019slum, owusu2024towards, persello2020towards, owusu2021towards, verma2019, fisher2022, cheng2022understanding, li2017unsupervised, hafner2022unsupervised, fan2022urban, luo2022urban} were retained and included in the final evaluation, highlighting the growing yet still nascent exploration of this field. This approach ensures that the findings are robust, laying a solid foundation for future inquiries into deep learning applications in mapping slums. The PRISMA flow diagram was used to visually depict the process of paper selection. The breakdown of the items included and excluded from the selection process is provided in Figure~\ref{fig:datasearch}. Our findings indicate a substantial potential for growth in the application of deep learning to slum mapping. This meta-analysis underscores the diversity in regional studies and data sources and the evolution of deep learning approaches in this domain. These insights can help improve urban planning and policy-making, directing resources more effectively to improve living conditions in slum areas.

Understanding the distribution of articles throughout different journals offers valuable information about the predominant literature sources within this field. Table ~\ref{tab:publications} illustrates the number of articles selected from different publications. The methodology in this study guarantees the reliability and accuracy of the findings and insights obtained from the review. These findings represent the current advancements in the application of deep learning techniques in mapping slum areas.

\begin{table}[ht]
\centering
\caption{Distribution of the selected articles across different journals}
\footnotesize
\begin{tabular}{p{5.5cm} c}
\hline
\textbf{Journal} & \textbf{Publications} \\
\hline
Sustainable Cities and Society & 1 \\
Photogrammetric Engineering and Remote \newline Sensing & 1 \\
Sustainability (Switzerland) & 1 \\
Habitat International & 1 \\
International Journal of Electrical and Computer Engineering & 1 \\
ISPRS Journal of Photogrammetry and Remote Sensing & 1 \\
IEEE Journal of Selected Topics in Applied Earth Observations and Remote Sensing & 1 \\
CEUR Workshop Proceedings & 1 \\
IEEE International Conference on Data Science and Advanced Analytics & 1 \\
ISPRS International Journal of Geo-Information & 1 \\
Sensors (Switzerland) & 1 \\
Cities & 1 \\
Neural Computing and Applications & 1 \\
Landscape and Urban Planning & 1 \\
International Archives of the Photogrammetry, Remote Sensing and Spatial Information \newline Sciences & 1 \\
npj Urban Sustainability & 1 \\
Remote Sensing Applications: Society and \newline Environment & 1 \\
International Journal of Applied Earth \newline Observation and Geoinformation & 2 \\
International Geoscience and Remote Sensing Symposium (IGARSS) & 2 \\
IEEE Transactions on Geoscience and Remote Sensing & 2 \\
Remote Sensing of Environment & 2 \\
IEEE Geoscience and Remote Sensing Letters & 2 \\
Scientific African & 2 \\
Joint Urban Remote Sensing Event, JURSE & 3 \\
Computers, Environment and Urban Systems & 4 \\
Remote Sensing & 4 \\
\hline
\end{tabular}

\label{tab:publications}
\end{table}

\subsection{Points of Focus}

Figure ~\ref{fig:datasearch} displays the total number of papers chosen for this meta-analysis study. This study aims to examine deep learning techniques in these papers in the context of slum mapping. Our analysis aims to address the following questions:

\begin{enumerate}
\item \textbf{What are the regions of study?} - We aim to identify the regions where these papers have trained and tested their models. This is important because slums may have very different characteristics in different parts of the world.
\item \textbf{What are the data sources and the datasets used?} - Since there are no well-known, annotated, and standardized datasets available for the task of slum mapping, the choice of the data source is an important decision for any comprehensive investigation of this task.
\item \textbf{How are the training and testing data prepared for the model?} -  In addition to the dataset itself, it is important to understand the pre-processing, data augmentations, imbalance corrections, and any other methodologies employed to adequately prepare the data for modeling.
\item \textbf{What is the structure of the predictive model?} - This delves into the technical details of the modeling process, including the architectures employed, loss functions, optimizers, and hyper-parameters.
\item \textbf{What are the different metrics involved in performance evaluation?} - The evaluation metrics play a pivotal role in understanding the efficacy and dependability of the models. Furthermore, the choice of metrics is vital to answering the specific questions being addressed by a work.
\end{enumerate}

\section{Analyzing the Points of Focus}
Now, we take a detailed look at each of the points of focus identified above.

\subsection{Study Regions}

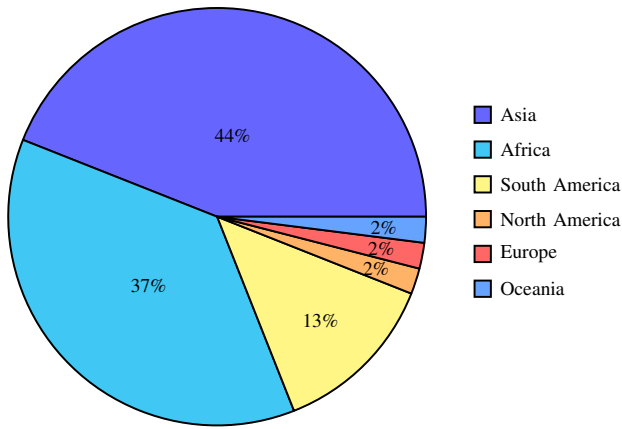
\begin{figure}[!ht]
    \centering
    \resizebox{\columnwidth}{!}{ 
        \begin{tikzpicture}
        \pie[
            radius=3,
            text=legend,
            font=\footnotesize
        ]{44/Asia,
        37/Africa,
        13/South America,
        2/North America,
        2/Europe,
        2/Oceania}
        \end{tikzpicture}
    }
    \caption{Continent-wise distribution of study regions}
    \label{fig:sa1}
\end{figure}

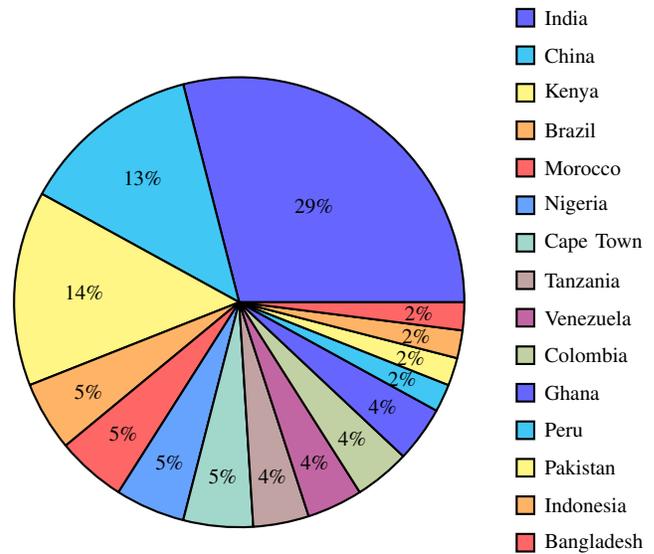
\begin{figure}[!ht]
    \centering
    \resizebox{\columnwidth}{!}{ 
        \begin{tikzpicture}
        \pie[
            radius=3,
            text=legend,
            font=\footnotesize
        ]{
            29/India,
            13/China,
            14/Kenya,
            5/Brazil,
            5/Morocco,
            5/Nigeria,
            5/Cape Town,
            4/Tanzania,
            4/Venezuela,
            4/Colombia,
            4/Ghana,
            2/Peru,
            2/Pakistan,
            2/Indonesia,
            2/Bangladesh
        }
        \end{tikzpicture}
    }
    \caption{Countries from where study regions are frequently selected}
    \label{fig:sa2}
\end{figure}

Here, we examine the geographical distribution of study regions that have been focused on the detection of slums using deep learning techniques. We conducted an analysis at both the continental and country-specific scales. Based on the analysis shown in Figure ~\ref{fig:sa1}, it is apparent that Asia, followed by Africa, was the primary region of interest for such studies. As demonstrated in Figure ~\ref{fig:sa2}, India leads the chart with 29\% of the research, followed by China and Kenya. The investigations on slums in India mostly focus on regions like Mumbai and Bangalore \cite{wang2019deprivation, dabra2023, ajami2019identifying, ansari2020, stark2020, wurm2019, stark2019slum, verma2019, fisher2022}. Shenzhen (\cite{chen2022hierarchical, huang2023comprehensive}) has emerged as a significant center of attention in China. Numerous African nations \cite{wahbi2023deep, mboga2017DL, abascal2022identifying}, namely Kenya, Morocco, and Tanzania, have also been the focus of these studies, with substantial research in Nairobi.

\subsection{Data Sources}

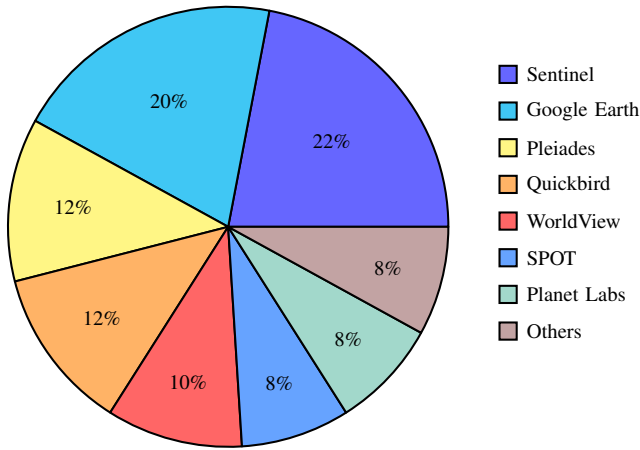
\begin{figure}[!ht]
  \centering
  \resizebox{\columnwidth}{!}{ 
    \begin{tikzpicture}
      \pie[
        radius=3,
        text=legend,
        font=\footnotesize
      ]{
        22/Sentinel,  
        20/Google Earth,
        12/Pleiades,
        12/Quickbird,
        10/WorldView,
        8/SPOT,
        8/Planet Labs,
        8/Others
      }
    \end{tikzpicture}
  }
  \caption{Dominant Satellite Platforms}
  \label{fig:datasources}
\end{figure}

The Deep Learning model requires high-quality annotated data for training and evaluation. Figure ~\ref{fig:datasources} illustrates the distribution of data sources used in deep learning-based slum mapping studies. Sentinel satellites are the most commonly used data sources, accounting 22\% of the studies, followed by Google Earth imagery at 20\%. Pleiades and Quickbird comprise 12\%, while WorldView 2 \& 3 account for 10\%. SPOT 5, 6, \& 7 and Planet Labs represent 8\% of the data sources. Other sources, such as PlanetScope, GeoEye, and TerraSAR-X, comprise the remaining 8\%.

This variety of data sources reflects the range of resolutions and spectral capabilities that researchers require for different aspects of slum mapping. The detailed imagery from providers like DigitalGlobe (e.g., Quickbird, WorldView) is crucial for fine-grained analysis and feature extraction within slum areas. QuickBird was a VHR commercial EO satellite launched in 2001 and decommissioned in 2015. It provided imagery at 60 cm in panchromatic (PAN) mode and 2.4 m in multispectral (MS) mode, capturing data in the blue, green, red, and near-infrared spectrum. QuickBird’s capabilities made it one of the leading sources for urban planning, environmental monitoring, and mapping services, offering detailed views that can be used for slum mapping and land-use studies \cite{wurm2019, li2017unsupervised}. The WorldView series are commercial EO satellites operated by Maxar Technologies. It provides PAN imagery at 0.46 m and eight-band MS imagery at 1.84 m, using four commonly used colors (red, green, blue, and near-infrared 1) and four spectral bands (coastal, yellow, red edge, and near-infrared 2). WorldView-3's 31 cm PAN resolution and MS bands—short-wave infrared (SWIR) and CAVIS (Cloud, Aerosol, Vapour, Ice, Snow)—help identify materials and terrain types. This makes WorldView satellites suitable for detailed mapping, including urban analysis and slum detection \cite{abascal2024ai, abascal2022identifying, ansari2020}. On the other hand, Sentinel satellites are advantageous for large-scale, regional studies. The Sentinel satellites are part of the Copernicus Programme, coordinated by the European Space Agency (ESA), providing a comprehensive EO system. Sentinel-1 provides radar imagery, ideal for monitoring urban expansion, land surface for motion risks, and infrastructure stability. Sentinel-2's 10 m resolution allows for comprehensive images of large areas, which is vital for monitoring land changes and vegetation. It is also extensively used for slum mapping \cite{raj2023mapping, owusu2024towards, fisher2022} as it is openly available. Google Earth provides accessible data that is frequently updated, which can be beneficial for time-series analyses and monitoring changes over time.

The usage of such diverse data sources underscores the adaptability of deep learning methods to handle different spatial resolutions and spectral characteristics critical for accurate slum detection and mapping. This distribution suggests that a combination of different data sources can offer a comprehensive understanding of slums. High-quality and varied data sources are essential for developing robust deep-learning models to achieve effective urban management and policy-making.

\subsection{Data Preparation}

Data preparation for deep learning applications in slum detection is a multi-step procedure designed to handle the complexities of urban informal settlements, as seen in satellite imagery. In this section, we present the methods employed in pre-processing, preparation, and augmentation, with a focus on their methodological rationale and contextual significance in slum mapping.

\begin{itemize}
    \item \textbf{Pre-processing}: It plays a crucial role in preparing remote sensing imagery for feature extraction and model training. This involves adjusting image attributes at the pixel or spectral level to better capture the physical characteristics of slum environments. For instance, spectral descriptors \cite{lu2024geoscience, chen2022hierarchical} are created from high-resolution imagery to capture the unique material signatures within slum environments. GLCM measures delineate the textural characteristics of slums, and structural patterns are retrieved to distinguish land features. Image enhancement and atmospheric correction are used to obtain surface reflectance values needed to train models to accurately interpret real-world conditions \cite{hafner2022unsupervised, pan2020deep}. Pan-sharpening methods are applied to MS bands to increase the clarity and detail of the imagery \cite{debray2019detection}, an essential step for resolving the fine-grained spatial structures prevalent in slums. Specialized software tools such as ArcGIS, which facilitates complex spatial analysis, and ENVI, which aids in the processing of geospatial data, are essential for enhancing and refining these images \cite{najmi2022integrating, pan2020deep}.
    \item \textbf{Data Cleaning and Integration}: Data cleaning involves de-skewing images to correct any tilt and cropping to focus on the areas of interest, removing irrelevant information that could confound the learning process \cite{bardhan2022geoinfuse}. Integrating datasets from diverse sources is a pivotal task as it allows for the consolidation of data, providing models with a comprehensive view of the urban landscape. Furthermore, binary masking delineates areas, while normalization standardizes data values, ensuring dataset consistency and comparability. 
    \item \textbf{Data Augmentation}: It plays a crucial role in augmenting the diversity and quantity of training data, thereby enhancing the accuracy and generalization capabilities of the model. In the context of slum mapping, where satellite imagery exhibits varied and complex urban textures, augmentation helps models adapt to the different visual characteristics of slums. Several key data augmentation techniques include random cropping and image tiling, which are used to focus on different parts of an image, ensuring the model can recognize slums from partial views. This is particularly useful when working with large areas, allowing the model to focus on detailed segments of the imagery, typically in sizes of 256x256 or 512x512 pixels. Flipping and rotation are also extensively used. Brightness adjustments and noise addition further enhance the feature set, ensuring that the model is robust. These transformations allow the model to learn generalized features that are resilient to changes in orientation, scale, and translation, a critical aspect for slum identification in heterogeneous urban landscapes \cite{abascal2024ai, abascal2022identifying, hafner2022unsupervised}. Affine transformations, which include minimal shearing and stretching, help the model generalize across different sensor angles and perspectives, mimicking real-world variations. To ensure effective learning, datasets are typically divided into training and testing subsets, often using an 80-20 or 70-30 split. Independent shuffling of the datasets is crucial to avoid biases related to the data, ensuring the model learns to generalize from a representative sample of urban landscapes. 
    \item \textbf{Ground Truth Label Generation}: Ground truth labels and masks are essential in these studies, as they act as reference data for training and evaluating models. Several studies used manual annotation of satellite images \cite{wahbi2023deep, huang2023comprehensive, abascal2022identifying, owusu2024towards}. This method entails the involvement of human experts who identify and demarcate urban villages, slum regions, and areas of deprivation. This process of manual annotation ensures accurate and reliable ground truth labels. Multiple studies integrated a variety of data sources, including remote sensing data, street view images, and social sensing data \cite{chen2022hierarchical, najmi2022integrating, fan2022multilevel}. The ground truth labels were created by combining these various data types, allowing for more thorough and precise model training. This label-generating process is enriched with expert knowledge as it offers valuable perspectives on distinct characteristics in urban regions, aiding in the creation of reliable labels. 
\end{itemize}

\begin{table*}[!h]
\centering
\caption{Data Preparation Techniques for Deep Learning in Slum Identification.}
\label{tab:dataprep}
\begin{tabular}{|l|p{7cm}|l|}
\hline
\textbf{Category} & \textbf{Techniques} \\
\hline
Pre-processing Techniques & Spectral descriptors extraction, GLCM metrics computation, structural patterns extraction, image enhancement, atmospheric adjustments, pan-sharpening of multispectral bands \\
\hline
Data Cleaning and Integration & De-skewing of images, cropping to focus areas of interest, data merging from various sources, normalization of data values\\
\hline
Data Augmentation Techniques & Random cropping, flipping and rotation, brightness enhancements, shearing and stretching \\
\hline
Dataset Preparation & Segmentation into training and testing, image tiling, and independent shuffling of datasets \\
\hline
Ground Truth Label Generation & Manual annotation, integration of diverse data for label generation with expert involvement \\
\hline
\end{tabular}
\end{table*}

The selected studies demonstrate a wide array of data preparation techniques (see table ~\ref{tab:dataprep}), indicative of the complex nature of slum environments in satellite imagery. The data preparation stage sets the stage for effective model training, essential to accurately map slums. Through meticulous pre-processing, cleaning, augmentation, and label generation, the groundwork for deep learning models to learn from and adapt to the complexity inherent in slums is established. The findings show that models must be trained on high-quality, representative datasets with extensive data preparation. Practitioners should focus on robust pre-processing to improve model accuracy. For policymakers, this underlines the importance of data collection and preparation standards, ensuring that efforts in slum mapping yield useful and reliable information for decision-making.

\subsection{Modeling Approaches} 

The training process plays a crucial role in implementing deep learning models for slum detection. Here, we outline the approaches used in training, encompassing network designs, frameworks, optimizers, and loss functions, among other factors.

\subsubsection{Network Topologies}

The performance of deep learning models in slum detection depends on choosing suitable network topologies based on the datasets described in the previous section. These models must not only accurately represent the complexities of urban landscapes but also be able to adjust to the inherent variability found in slum areas. A diverse range of architectural models has been developed, including semantic segmentation using variations of the U-Net \cite{wahbi2023deep, lu2024geoscience, wang2019deprivation, raj2023mapping, hafner2022unsupervised}. These models use deep learning techniques to accurately identify and outline the complex and irregular patterns found in slums. Additionally, hybrid models like GASlumNet \cite{lu2024geoscience} combine the strengths of well-established networks like ConvNeXt to enhance performance. Another method involves using pre-trained models that leverage the knowledge gained from other large image datasets. This allows for the necessary adjustments to be made in order to accurately map slum areas. The refinement of such models is enhanced through unique architectures and multimodal networks \cite{fan2022fine, fan2022urban}, highlighting the intricate nature of slum detection. This is evident in networks that incorporate transformer-based fusion approaches \cite{fan2022fine, fan2022urban}, which indicates a move towards using a variety of data sources. These deep learning models are carefully curated to turn raw data into practical insights about network topology.

\subsubsection{Loss Functions and Optimization Algorithms}

\begin{figure}[!hb]
\centering
  \resizebox{\columnwidth}{!}{ 
    \begin{tikzpicture}
      \pie[
        radius=3,
        text=legend,
        font=\footnotesize
      ]{
        58/Adam,  
        29/SGD \& SGD with momentum,
        13/Others
      }
    \end{tikzpicture}
  }
  \caption{Distribution of Optimizers Employed in Slum Detection Studies, with Adam emerging as the predominant choice}
  \label{fig:optimzers}
\end{figure}
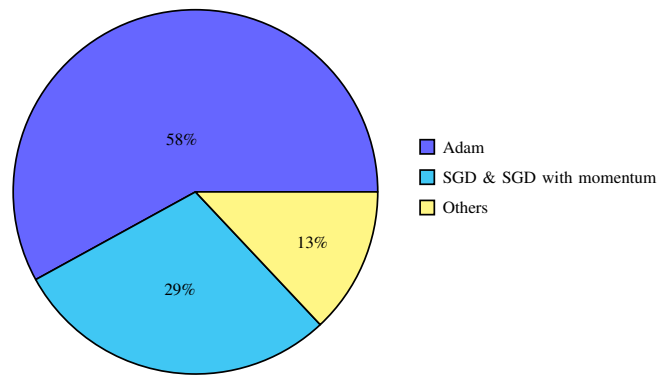

Deep learning for slum identification has used a variety of loss functions and optimization techniques to address the challenges posed by the complexity of urban informality. The Adam optimizer, as shown in Figure ~\ref{fig:optimzers}, has been a popular choice because of its ability to adjust the learning rate based on the significant variability found in slums. This optimizer, typically initialized with learning rates of 0.001, has played a pivotal role in various research, enabling consistent convergence and dependable performance.

The commonly used loss function in conjunction with Adam optimization is the weighted cross-entropy (WCE) loss. This loss function addresses class imbalance by allocating varying weights to classes according to their frequency. It is particularly important in slum identification, as the slum areas may only make up a small part of the cityscape yet are of great importance. Simultaneously, the Dice loss function is frequently used to segment slum regions, boosting model sensitivity and specificity by aligning predicted and ground-truth segments. A study also suggests combining Dice loss with WCE to make a hybrid loss function \cite{lu2024geoscience} to effectively deal with the intricacies of slum detection. Dice loss, which emphasizes spatial overlap, and WCE, which handles imbalanced datasets, create a comprehensive learning process.

The use of Adam and hybrid loss functions is the prevailing trend in deep learning methodologies for slum detection. However, the optimization algorithm and loss functions can be meticulously adjusted to tackle the unique requirements of the task. These methods train deep learning models to not only distinguish between slum and non-slum areas but also to enhance their understanding of the varying levels of informality within the urban landscape. This is crucial for accurate mapping and subsequent interventions.

\subsubsection{Deep Learning Framework, Hyperparameters, and Regularization}

\begin{figure}[hb]
  \centering
  \resizebox{\columnwidth}{!}{ 
    \begin{tikzpicture}
      \pie[
        radius=3,
        text=legend,
        font=\footnotesize
      ]{
        37/Keras,  
        33/Tensorflow,
        19/PyTorch,
        11/Others
      }
    \end{tikzpicture}
  }
  \caption{Preference of Deep Learning Frameworks in Slum Detection Research, showcasing TensorFlow's dominance}
  \label{fig:framework}
\end{figure}
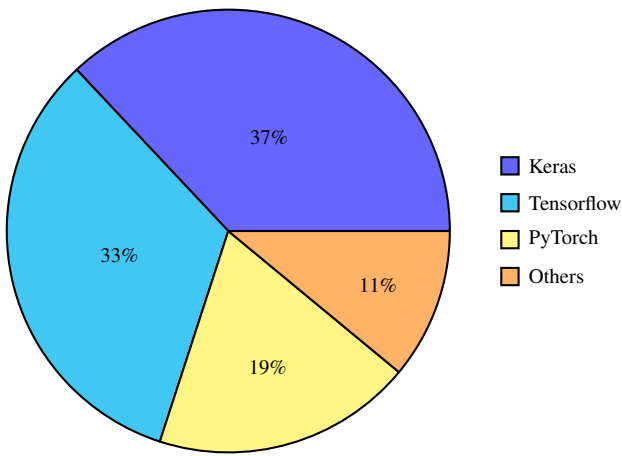

Deep learning frameworks, hyperparameter optimization, and regularisation techniques are crucial for improving the model performance for slum identification. TensorFlow and Keras are the dominant frameworks in this field (see Figure ~\ref{fig:framework}). Their user-friendly interface, extensive documentation, and strong community support make them ideal for academic and practical research. PyTorch, TensorFlow, Keras, and Theano are known for their adaptability and comprehensive libraries, enabling various network designs and training methods. PyTorch's adaptability has been advantageous for developing topologies like U-Net in conjunction with ConvNeXt \cite{lu2024geoscience}, while Keras and TensorFlow are popular for their user-friendly APIs and simplicity in applying transfer learning \cite{el2024new, verma2019}. These frameworks enable the creation of various deep learning models with intricate topologies and offer smooth support for computing, which is essential for handling large remote sensing datasets.

Hyperparameter optimization is crucial for model performance as it directly impacts the model's convergence rate and overall performance. Learning rates commonly commence at $10^{-3}$ or lower, whereas weight decay parameters are typically within the range of $10^{-4}$ to $10^{-5}$ to guarantee robust learning without encountering overfitting. Several techniques, such as the Cosine learning rate decay and warm-up procedures \cite{huang2023comprehensive}, are used to optimize the learning process dynamically. This meticulous calibration enables models to successfully navigate the complex features of slums. 

Regularization techniques are commonly employed to mitigate overfitting, a prevalent issue when training models on high-dimensional data. L2 regularization and Dropout ensure that every neuron contributes to learning by adapting to the model's structure and training patterns. The use of Xavier and He initialization, as well as BatchNorm and Dropout, are used to make the model better at applying what it has learned from training data to situations it has never seen before in the real world.

These strategies improve model training using initialization and regularization techniques, optimized learning rates, and powerful deep-learning frameworks. The meticulous handling of hyperparameters and regularisation at a granular level highlights the need for a nuanced strategy to tackle the intricacies of slums in satellite imagery. With advanced training methods, deep learning models have improved their ability to identify and categorize slum regions. 

\subsection{Metrics Used}

Deep learning for slum detection involves evaluating models across a range of metrics that capture the characteristics of urban informality. These metrics interlace diverse architectural choices and preprocessing strategies discussed in previous sections to determine model applicability. The common metrics of accuracy, precision, recall, and the F1 score form the cornerstone of model evaluation. The true positive rate (TPR) and false positive rate (FPR) for ROC curve analysis are additions to these, providing insights into the trade-offs between sensitivity and specificity. Meanwhile, the Jaccard Index serves as an indispensable metric, especially pertinent for segmentation tasks, gauging the precision of overlap between the predicted slum areas and ground truth.

The overall accuracy (OA) shows how well the model fits the data, while the Intersection over Union (IoU) or Jaccard Index looks at how well the segments overlap, which is important given the slums' spatial interweaving. Precision and recall, along with the harmonizing F1-score, dissect the models' ability to assess urban complexity. The Kappa coefficient tempers these measures, adding a layer of statistical robustness to the analysis. In class-wise accuracy, the spotlight shifts to models' capabilities in differentiating slum typologies. The producer's and user's accuracy lend insight into the models' reliability. The Area Under the Receiver Operating Characteristic (AUROC) shows how well a binary classification works, while the Area Under the Precision-Recall Curve (AUPRC) addresses the class imbalance in urban datasets. Visual inspections add a qualitative dimension to this quantitative array, ensuring that models hold up to statistical scrutiny and the human eye—a crucial check for models for policy and urban planning.

The evaluation metrics employed in the reviewed studies reveal a multi-faceted approach to model assessment. OA, IoU, precision (P), recall (R), and F1-score (F1) are among the most commonly used metrics, providing a comprehensive view of model performance across different dimensions. The emphasis on these metrics confirms their importance in validating the reliability of deep learning models in slum mapping. Multiple metrics allow researchers to assess the accuracy, applicability, and fairness of the models across various urban contexts. The precision generally falls within 87\% to 95\%, demonstrating the models' capacity to detect urban characteristics. Meanwhile, recall rates typically show significantly lower values, indicating the difficulty of accurately recording all relevant instances. The accuracy of the models reaches high values, typically above 90\%, which highlights their overall usefulness. F1-score, Jaccard Index, Kappa (K), and mean intersection over union (mIoU) frequently surpass 0.8, reflecting a robust balance between precision and recall. This comprehensive evaluation suggests that deep learning and remote sensing data yield reliable results in slum mapping for urban planning and development. 

By engaging with this elaborate suite of metrics, researchers ensure that their models are theoretically sound and practically viable. For practitioners, these metrics underscore the importance of rigorous validation processes, and for policymakers, they spotlight the necessity of supporting initiatives that emphasize a detailed approach to model assessment. This comprehensive metrics analysis underscores the transition from precision engineering in model development to precision efficacy in real-world applications, enabling data-driven urban planning and policy-making.

\section{Deep Learning Solutions in GIS Software}

Deep learning applications in remote sensing have enhanced the analytical capabilities of Geographic Information System (GIS) software, simultaneously improving accessibility and user-friendliness. This advancement is characterized by the incorporation of image classification tools coupled with user-friendly graphical user interfaces (GUIs), making these computational tools more accessible to individuals with rudimentary programming knowledge.

Table \ref{tab:DLsolutions} provides an overview of the deep learning functionalities available in commercial and open-source GIS software. This underscores the industry’s commitment to promoting robust, user-friendly deep learning tools that facilitate the broader adoption of CNNs among a diverse user base. Open-source tools like QGIS and Orfeo ToolBox offer flexibility and community support, appealing to users who favor transparency and customization. However, proprietary software such as ERDAS Imagine and ArcGIS Pro provide integrated, out-of-the-box solutions with enterprise-level support, which might be preferable for users seeking a more guided experience. 

\begin{table*}[htb]
\centering
\caption{Overview of deep learning functionalities in prevalent GIS software.}
\label{tab:DLsolutions}
\begin{tabular}{|l|p{6cm}|l|}
\hline
\textbf{Software} & \textbf{Deep Learning Functionality} & \textbf{Licensing Model} \\
\hline
QGIS & Supports ONNX-based inference for tasks like segmentation and detection through the Deepness module. & Open-source \\
\hline
Orfeo ToolBox & Features the OTBTF module, providing a deep learning framework specifically designed for processing satellite imagery. & Open-source \\
\hline
ERDAS Imagine & Offers tools for image classification and feature extraction, including capabilities for semantic segmentation and object recognition. & Proprietary \\
\hline
eCognition & Facilitates the creation, training, and deployment of CNNs utilizing the TensorFlow platform, tailored for intricate image analysis tasks. & Proprietary \\
\hline
ENVI & Includes a specialized module for training deep learning models using TensorFlow, enhancing its utility in satellite and aerial image analysis. & Proprietary \\
\hline
ArcGIS Pro & Supports deep learning analyses for both images and point clouds, incorporating architectures such as SSD and U-Net. & Proprietary \\
\hline
GRASS GIS & Integrates with Python libraries like TensorFlow or PyTorch for spatial data analysis, offering capabilities for complex spatial modeling. & Open-source \\
\hline
PostgreSQL with PostGIS & Facilitates the use of deep learning via extensions like PL/Python, allowing for sophisticated spatial queries and data analysis within a database framework. & Open-source \\
\hline
\end{tabular}
\end{table*}

In addition to these platforms, the FOSS4G (Free and Open Source Software for Geospatial) ecosystem democratizes advanced geospatial processing tools. It provides several deep learning tools and modules, including Python-integrated spatial analysis processing packages.  These open-source technologies encourage innovation and collaboration in the geospatial community.

As the geospatial field looks to the future, the integration of deep learning tools in urban planning and policy-making frameworks presents opportunities and challenges. These tools offer the potential to drive significant advancements in urban management, but they must be tailored to varied urban settings. These advanced analytical tools can lead to more informed decision-making and contribute to urban sustainability. However, the transformative power of deep learning in GIS must be managed with a sense of responsibility, ensuring that its application is beneficial, ethical, and conducive to the public good.

\section{Discussion}

The geographical scope and methodological breadth of the reviewed research show that the integration of deep learning in the study of urban slums has increased in recent years. This paper has dissected the layers of this integration, presenting insights into regional focuses, data diversity, model architectures, and the complex socio-economic tapestries these slums present.

\subsection{Regional Focus and Implications}

A significant concentration of research has been on slums in Asia and Africa. The regions of Mumbai, Shenzhen, and Nairobi have each emerged as focal points in the application of deep learning for urban slum study, highlighting the global concern and interest in improving living conditions through enhanced urban planning and policy-making. Each of these regions represents unique facets of the global slum phenomenon. 
\begin{itemize}
    \item \textbf{Mumbai} epitomizes the contrasts of thriving economic growth alongside sprawling slums. It is home to some of the world's largest informal settlements, like Dharavi. Researchers have targeted Mumbai due to its significant slum population, which presents many challenges, from congestion to limited access to basic services, among others. The diverse texture of Mumbai's slums and their density make it an ideal case study for deploying and testing deep learning models. To effectively map dense and diverse slum regions, models must combine spectral and textural information, as demonstrated by the remarkable performance metrics of GASlumNet \cite{lu2024geoscience}. Studies employing modified CNN models, like VGG16-UNet and MobileNetV2-UNet, indicate a focused effort to identify not just the slum areas but also essential features within them, such as green cover and open spaces \cite{dabra2023}. These studies achieve high precision and recall, signifying the models’ accuracy in classifying fine-grained details. Using pre-trained networks like VGGNet, ResNet, and DenseNet in GeoInFuse \cite{bardhan2022geoinfuse} shows that existing architectures are being improved by combining multiple channels of data to find different types of urban forms. This could have profound implications for distinguishing between formal and informal urban areas, which is crucial for effective urban planning.
    \item \textbf{Shenzhen}, one of China's most populous cities, has transformed into a megacity and hub of technology and innovation. Rapid urbanization and the migration of millions into the city have led to the growth of informal settlements, often overshadowed by the city's global economic status. Studies such as the hierarchical recognition framework for urban villages \cite{chen2022hierarchical} combine remote and social sensing data to address the challenge of detecting urban sprawls. Integration of street-level imagery in the Vision-LSTM module \cite{huang2023comprehensive} enhances human-centric analysis of urban sprawl. The application of a transformer-based multimodal fusion network (UisNet) \cite{fan2022fine} and a multilevel spatial-channel feature fusion network (FusionMixer) \cite{fan2022multilevel} indicates a cutting-edge direction in slum research. These networks suggest that using multimodal data can lead to a more nuanced representation of urban spaces, which could help with the accuracy of classifying different types of urban forms.
    \item \textbf{Nairobi's} slums are among the most studied in Africa. These settlements are characterized by high population densities, inadequate infrastructure, and services. It has grappled with housing inadequacies, a legacy of its colonial past that continues to challenge its urban landscape. The complexity of Nairobi's slums and socio-economic issues pose unique challenges to slum mapping. Nairobi provides a stark representation of the quintessential African urban slum, rich in community life but facing significant development challenges. The studies here have not only focused on the mapping of slums but also on predicting the perceptions of deprivation among citizens \cite{abascal2024ai}, reflecting a multi-dimensional approach to understanding slum environments. The emphasis on models that map deprivation from citizen votes in Nairobi represents a shift towards more inclusive, community-focused urban research. The "EO + Morphometrics" methodology \cite{wang2023eo+} highlights the significance of urban morphology in satellite data interpretation, enabling more accurate and reproducible urban pattern mapping.
\end{itemize}

The application of deep learning models across these regions is not just a technological pursuit but also an attempt to incorporate socioeconomic factors. The cross-regional study of these areas reveals intriguing patterns and common challenges, emphasizing the need for technically robust and socioeconomically responsive models for slum inhabitants. Mumbai and Shenzhen focus on high-resolution imaging for detailed urban analysis, and Nairobi uses a combination of VHR imagery and socio-economic data, demonstrating the importance of community-focused research in urban planning. Tables \ref{tab:mumbai_results}, \ref{tab:shenzhen_results}, and \ref{tab:nairobi_results} summarize the key findings from the predominant studies across these regions, reflecting a range of methodologies and satellite data used. The diversity in data sources, from satellites like WorldView and Pleiades to Sentinel-2, underlines the balance required between detailed imagery and spectral analysis. The comparison of model performance metrics across regions reveals variations in focus and outcomes. Mumbai shows high accuracy in slum segmentation, while Shenzhen's models excel in integrating diverse data types for urban analysis. Nairobi's approach highlights the integration of deep learning with citizen science to enhance the social relevance of technological interventions. This geographical skew suggests an urgent need for solutions tailored to the specific challenges of rapidly urbanizing regions. It also raises concerns about data availability and representativeness, urging researchers to consider local context and physical variances in their methods. The discussion of these aspects in the context of each study provides a comprehensive picture of deep learning applications in slum mapping and sets the stage for future research focused on technological advancements as well as socio-economic impacts.

\begin{table*}[htb]
\centering
\caption{Key Studies in Deep Learning for Mumbai Slum Mapping.}
\label{tab:mumbai_results}
\begin{tabular}{|m{1.3cm}|m{6.5cm}|m{2.5cm}|m{3cm}|m{2.5cm}|}
\hline
\textbf{Year (Study Reference)} & \textbf{Study Description} &\textbf{Satellite used \newline (Resolution in m)}  & \textbf{Model employed}  & \textbf{Performance \newline metrics}  \\ 
\hline
2024 \cite{lu2024geoscience} & This study introduced GASlumNet, a geoscience-aware network mapping accurate slum. It aimed to incorporate geoscientific knowledge, including spectral features and textural information. & Spectrum-01 (5m) and Sentinel-2 (10m) & GASlumNet, a combination of UNet and ConvNeXt & OA: 94.89\% \newline P: 0.7613\newline R: 0.7985\newline IoU: 0.6386         \\ 
\hline
2023 \cite{dabra2023} & This study trained three modified CNN models on manually labeled high-resolution satellite imagery to identify green cover and open spaces within informal settlements. & Pleiades-1A (0.5m) & VGG16-UNet, MobileNetV2-UNet, and DeepLabV3+. & VGG16-Unet (best): \newline OA: 95.02\% \newline P: 0.9508 \newline R: 0.9502 \newline F1: 0.9502 \\                  
\hline
2022 \cite{bardhan2022geoinfuse} & This study developed an end-to-end, multi-channel supervised data fusion framework called GeoInFuse to classify different urban form typologies in heterogeneous hyper-dense cities. & LISS   IV (5.8m) with the Cartosat (1m) & Pre-trained networks such as VGGNet, ResNet, and DenseNet within the GeoInFuse framework. & GeoInFuse-ResNet (best): \newline   OA: 74.98\% \newline P: 0.739 \newline R: 0.750 \newline F1: 0.737  \\               
\hline
2020 \cite{ansari2020} & This study used contourlet transforms and U-net architecture to construct multiscale deep learning techniques for recognizing informal built-up areas. This strategy helps the model extract and use detailed textural and spatial characteristics.
. & Worldview-2 (2m) & U-net (modified), augmented with contourlet masks of different sizes at different scales.  & OA: 94.98\% \newline P: 0.9345 \newline R: 0.9101 \newline F1: 0.9221 \newline mIoU: 0.89  \\                      
\hline
2023 \cite{raj2023mapping} & This study evaluated the U-Net model for slum delineation and categorized it as slums or non-slums. & Sentinel-2 (10m)  & U-Net & OA: 98.2\% \newline P: 0.973 \newline R: 0.983 \newline F1: 0.978\\
\hline
2019 \cite{wurm2019} & This study explored the application of transfer learning in fully convolutional networks (FCNs) for the semantic segmentation of slums. It leverages pre-trained networks on very high-resolution optical satellite imagery and transfers this learning to different satellite data types. & QuickBird (0.5 m), Sentinel-2 (10 m), and TerraSAR-X (6m).& Fully Convolutional Network (FCN) based on the VGG19 architecture & Transfer Learning: Quickbird to Sentinel-2 (best): OA: 89.64\% \newline K: 0.85 \newline IoU: 0.8743\\  
\hline
2019 \cite{stark2019slum} & This study addressed the challenge of slum mapping in urban environments with highly imbalanced datasets using transfer learning in FCNs. & QuickBird (2.4m); pan-sharpened to 0.6m  & Fully Convolutional Network (FCN) based on the VGG19 architecture  & IoU: 0.72  \\
\hline
2019 \cite{verma2019} & This study investigated transfer learning with pre-trained convolutional networks for mapping urban slums using VHR and medium-resolution satellite imagery. & Pleiades-1A (0.50m) and Sentinel-2 (10m) & Used pre-trained convolutional network (Inception-v3) & For VHR Imagery (best):  \newline OA:94.3\% \newline K: 0.704 \newline IoU: 0.583 \\ 
\hline
2022 \cite{fisher2022} & This study introduced an approach notable for quantifying pixel-wise uncertainty in predictions, enhancing the interpretability and reliability of slum mapping efforts. & Sentinel-2 (10m) & U-Net model with Monte Carlo Dropout (MCD) for uncertainty quantification  & AUPRC   score: 0.74 \newline Uncertainty value: 9.7 x   10-9\\
\hline
\end{tabular}
\end{table*}

\begin{table*}[htb]
\centering
\caption{Deep learning studies on mapping slum settlements in Shenzhen.}
\label{tab:shenzhen_results}
\begin{tabular}
{|m{1.3cm}|m{6.5cm}|m{2.5cm}|m{3cm}|m{2.5cm}|}
\hline
\textbf{Year (Study Reference)} & \textbf{Study Description} &\textbf{Satellite used \newline (Resolution in m)}  & \textbf{Model employed}  & \textbf{Performance \newline metrics}  \\ 
\hline
        2022 \cite{chen2022hierarchical} & This study used remote and social sensing data to create a hierarchical recognition framework for fine-grained urban villages (UVs). The methodology improves the accuracy and interpretability of UV mapping by using a multi-scale approach that mimics human cognitive processes. &  SPOT-5 (2.5 m) & A combination of machine learning classifiers and geospatial data analysis. & OA: 96.23\% \newline K: 0.920 \\ \hline
        2023 \cite{huang2023comprehensive} & This study presented a deep learning module that integrates varying numbers of street-level images to represent the characteristics of an urban spatial unit. This approach enables a comprehensive and representative understanding of urban spaces.  & Collected from Google Earth (0.6m) & Vision-LSTM module: comprises a Convolutional Neural Network (CNN) with shared weights and a Recurrent Neural Network (RNN).  & OA: 91.6\% \newline F1: 0.773 \newline K: 0.720 \\ \hline
        2022 \cite{fan2022fine} & This study focused on improving the semantic segmentation of urban informal settlements (UISs) using a transformer-based multimodal fusion network. The method leverages remote sensing images and building polygon data to achieve fine-scale mapping of UISs. & Collected from Google Earth (1.19m) & UisNet (transformer-based multimodal fusion network) & OA: 94.80\% \newline mIoU: 0.8551 \\ \hline
        2022 \cite{fan2022multilevel} & This study introduced a multilevel spatial-channel feature fusion network to improve UV classification by integrating satellite and street view images. The method aims to overcome the limitations of single-modality data by fusing features at multiple spatial and channel levels. & Collected from Google Earth (1.19m) & FusionMixer (CNN-based feature extraction module) and a multilevel spatial-channel feature fusing layer. & OA: 94.43\% \newline K: 0.8757 \\ \hline
        2017 \cite{li2017unsupervised} & This study introduced an unsupervised deep learning method for detecting UVs. This approach uses a deep CNN to autonomously learn feature representations, which is particularly valuable given the scarcity of labeled data in remote sensing applications. & QuickBird (2.4m) & An unsupervised deep neural network comprising an unsupervised deep convolutional neural network and an unsupervised deep fully connected neural network. & OA: 98.55\% \newline K: 0.9623 \\ \hline
        2022 \cite{fan2022urban} & This study addressed the classification of UIS by integrating remote sensing images with time-series Tencent population density (TPD) data using a hybrid deep learning network, aiming to improve the accuracy of UIS classification by capturing both spatial features of geological objects and temporal features of human activity patterns. & Collected from Google Earth (1.19m) & STNet, which includes PDNet, ResMixer, and a Transformer-based spatial-temporal fusing layer. PDNet handles time-series TPD data, ResMixer processes VHR images, and the Transformer layer fuses these inputs. & OA: 88.58\% \newline K: 0.7716 \\ 
\hline
\end{tabular}
\end{table*}

\begin{table*}[htb]
\centering
\caption{Deep learning studies on Nairobi's urban slums.}
\label{tab:nairobi_results}
\begin{tabular}
{|m{1.3cm}|m{6.5cm}|m{2.5cm}|m{3cm}|m{2.5cm}|}
\hline
\textbf{Year (Study Reference)} & \textbf{Study Description} &\textbf{Satellite used \newline (Resolution in m)}  & \textbf{Model employed}  & \textbf{Performance \newline metrics}  \\ 
\hline
        2024 \cite{abascal2024ai} & This study developed methods for predicting citizens' perceptions of deprivation using satellite imagery, citizen science, and AI. It explores AI's ability to model perceptions of deprivation derived from votes by slum citizens, with deep learning outperforming conventional machine learning methods. These models aid in comprehending and predicting how locals perceive the physical aspects of urban deprivation. & WorldView-3(1.20 m) & 2 different Convolutional Neural Networks (CNN): VGG and DenseNet-121.  & DenseNet-121 pretrained RGB bands (best): \newline
        Max R$^2$: 0.841 \newline
        Min RMSE: 0.693 \newline
        Mean R$^2$: 0.801 \newline 
        Mean RMSE: 0.767 \newline
        Std R$^2$: 0.021 \newline 
        Std RMSE: 0.039\\ \hline
        2023 \cite{wang2023eo+} & This study presented "EO + Morphometrics," workflow to improve the scientific validity and reproducibility of urban pattern mapping using EO and urban morphometrics. This approach focuses on detecting physical urban elements from satellite imagery and then applying urban morphometrics to these elements to derive interpretable urban patterns. & Collected from Google Earth (0.6 to 1.2m) & U-Net with ResNet-50 as the encoder & Accuracy: 93.79\% \\ \hline
        2022 \cite{abascal2022identifying} & This study applied deep learning techniques and morphological analysis to identify degrees of deprivation in deprived urban areas (DUAs).  & WorldView-3(1.20 m) & U-Net architecture for semantic segmentation & F1: 0.84 \newline IoU: 0.73 \\ \hline
        2024 \cite{owusu2024towards} & This study proposed a scalable and transferable method to map deprived areas across cities and countries in a consistent manner.  & Sentinel-2 (10m) & Multi-Layer Perceptron (MLP), Random Forest (RF), Logistic Regression (LR) and Extreme Gradient Boosting (XGBoost). & MLP (best): \newline  F1: 0.79 \newline P: 0.98 \newline R: 0.60 \\ \hline
        2022 \cite{luo2022urban} & This study investigated the potential of VHR EO-based data to predict degrees of intra-urban deprivation. It involved a two-step workflow: characterizing multiple dimensions of deprivation using Principal Component Analysis (PCA) and predicting these degrees using a CNN. & SPOT-7 (1.5m) & CNN-based regression model & R$^2$: 0.6543 \newline  RMSE: 0.0582 \\ \hline
    \end{tabular}
\end{table*}

\subsection{Observations on Model Selection in Deep Learning for Slum Mapping}

Our meta-analysis, which reviews 40 scholarly articles from 2014 to 2024, highlights the application of deep learning in slum mapping. The meta-analysis scrutinized articles selected through a meticulous process, leveraging databases such as Web of Science, Scopus, and Science Direct. PRISMA protocols guided this selection to ensure the integrity and relevance of the findings. Our study revealed the region examined the most, varying data sources from VHR satellites like WorldView to Sentinel satellites, and a diverse array of deep learning applications tailored to these contexts. The diverse dataset preparation and model application underscore the adaptability required to manage the complexities of urban informal settings effectively. This meta-analysis further emphasizes the need for context-specific model selection. CNNs have performed well in places with complicated spatial features as they can handle large amounts of data effectively \cite{abascal2024ai, el2023building, persello2017deep, debray2019detection, mboga2017DL, ajami2019identifying}. Alternatively, U-Nets excel at segmenting dense slum regions where precise delineation of boundaries is essential \cite{wahbi2023deep, wang2019deprivation, ansari2020, raj2023mapping, hafner2022unsupervised}. Hybrid models, combining features from various architectures, are promising for combining data types, such as satellite imagery and urban data \cite{lu2024geoscience, huang2023comprehensive, stark2023detecting, fan2022fine, fan2022urban}. Furthermore, the Adam optimizer has been widely adopted due to its efficiency in handling sparse gradients and adaptability to different data modalities, making it suitable for the variable nature of slums in satellite imagery. The WCE loss is prevalent due to its effectiveness in dealing with class imbalances—a common issue in slum datasets where non-slum areas vastly outnumber slum areas. The incorporation of Dice loss in hybrid loss functions helps improve model performance on segmentation tasks by emphasizing spatial overlap, which is crucial for accurate slum boundary detection. This study also underscores the increasing integration of deep learning models with GIS, which enhances spatial analysis capabilities. This integration is vital for translating model outputs into actionable insights for urban planning and policy-making, emphasizing the need for models that can seamlessly operate within GIS platforms. Following our comprehensive review, several key observations regarding model selection emerged:
\begin{itemize}
    \item \textbf{Model Suitability Across Different Regions}: Specific models, notably CNNs, showed enhanced performance in regions with dense urban fabrics, where their ability to segment detailed slum features proves invaluable.
    \item \textbf{Impact of Data Quality on Model Choice}: The quality of data significantly influences model effectiveness, with higher-resolution imagery favoring complex models capable of detailed feature processing. In contrast, areas with lower-resolution data might benefit from simpler models that require less computational power.
    \item \textbf{Adaptability and Flexibility}: The robustness of models incorporating transfer learning or those pre-trained on diverse datasets highlights their suitability for adapting to new, varied slum environments.
    \item  \textbf{Technical Constraints and Resource Availability}: Model selection often hinges on available technical resources, with simpler models preferred in resource-constrained settings to ensure broader accessibility.
    \item \textbf{Future Directions in Model Development}: The integration of AI with GIS points to future research directions focusing on hybrid models that merge deep learning's analytical strength with GIS's spatial analysis capabilities.
\end{itemize}
This synthesis of our meta-analysis with focused observations provides a robust framework for enhancing the precision and impact of deep learning in slum mapping, guiding future urban planning and policy-making to be more data-driven and inclusive.

\subsection{Issues and possible solutions}

The use of deep learning techniques in slum mapping through the analysis of EO data presents significant potential for revolutionary outcomes. These strategies possess the capacity to significantly augment our comprehension and surveillance of informal settlements. Nevertheless, it is necessary to acknowledge and tackle certain challenges to achieve optimal results.

\begin{itemize}
    \item \textbf{Data Availability}: The effectiveness of using deep learning techniques for mapping slums is contingent upon extensive training data. The lack of localized data \cite{georganos2022census} during the initial stages might significantly compromise the dependability and precision of mapping solutions due to the varied characteristics of slums. Without robust and diverse training datasets, models cannot be expected to perform with high accuracy across various geographies. Policymakers and funding agencies must recognize the need for and invest in the collection of local data, which can significantly enhance the models' utility.
    \item \textbf{Dataset Bias}: The issue of dataset bias arises when considering the representation of slums in satellite and EO data. Slums, characterized by their informal and temporary nature, may not be sufficiently captured in these datasets despite the ongoing availability of such data. The existence of diverse slum areas across different locations \cite{stark2023detecting} presents a significant problem in acquiring the comprehensive dataset necessary for accurate training and detection. Governments, NGOs, and academia can collaborate to develop more representative datasets for more equitable and effective urban planning.
    \item \textbf{Explainability}: The concept of explainability in the context of slum mapping carries significant socio-economic and policy consequences. As these models inform decisions that affect urban populations, especially the marginalized, it's essential that the models' decision-making processes are transparent \cite{hall2022review}. This calls for interdisciplinary collaboration, where urban planners, data scientists, and social workers come together to ensure the models are not just effective but also just and fair.
    \item \textbf{Privacy and Ethical Concerns}: The application of deep learning in urban monitoring, particularly in the context of slum mapping, gives rise to privacy and ethical concerns \cite{kochupillai2022earth} that must be addressed. Improper use of data may result in unwarranted monitoring and harm vulnerable individuals. Ethical considerations \cite{owusu2021geo} extend beyond the realm of privacy. An ethical concern arises from the absence of informed consent among individuals residing in slums, who are frequently unaware of being subjected to technological research. The lack of transparency in their data collection, processing, and use gives rise to ethical concerns. Furthermore, these studies have the potential to stigmatize slums by spreading prejudices and disregarding their intricate socio-economic dynamics. The ethical ramifications of data interpretation in policymaking have similar significance. Urban planners who make decisions based on deep learning algorithms should give priority to the well-being and rights of residents to prevent the exacerbation of marginalization or displacement.
\end{itemize}

The necessity of extensive labeled training datasets for slum mapping is particularly evident in deep learning \cite{stark2024quantifying}. To ensure efficient model training, high-quality and diverse data that precisely depicts slum patterns and differences across geographical regions is essential. CNNs, known for their spatial recognition performance, depend on the amount and detail of training data, especially when dealing with intricate structures in slum areas. Due to the complex nature of urban slums, it is imperative for models to accurately identify and analyze minute details, which may require multiple layers within the network \cite{lumban2023comparison}, increasing the computational requirements. So, finding the right balance between computing efficiency and algorithmic depth becomes crucial. Overly complex networks might capture noise instead of meaningful patterns, potentially leading to overfitting and misclassification. The importance of deep learning models lies in their adaptability. The characteristics of slums in a city or country might exhibit substantial variations compared to those in other locations \cite{reviewkuffer, reviewron}. The applicability of models trained on a particular dataset, such as slums in Mumbai, to slums in Nairobi may be limited due to localized differences in structural characteristics, building materials, and spatial arrangement. The use of deep learning in slum mapping through the analysis of EO data holds significant potential. However, achieving this potential depends on effectively overcoming the inherent obstacles associated with this approach. The effective and responsible use of deep learning techniques for urban monitoring and planning necessitates a comprehensive comprehension of slums and the implementation of robust data techniques and transparent models. It is also imperative to establish and adhere to rigorous ethical standards for data collection \cite{lobo2023ethics}, analysis, and use to tackle these difficulties. The emphasis should be on prioritizing privacy and ensuring data security when establishing these standards. Involving local communities in the process of data collection guarantees that their perspectives are considered \cite{thomson2020need}, fostering an inclusive and ethical methodology. Ethicists, sociologists, community leaders, technologists, and urban planners might collaborate to gain a more comprehensive understanding of the ethical implications associated with the use of deep learning in urban monitoring.

The use of deep learning for slum mapping poses complex issues. One of the primary challenges lies in accurately defining and describing it \cite{reviewkuffer}. To overcome this challenge, it is crucial to integrate satellite imagery with street-level data. This will provide not only a perspective on individual slum structures but also on the broader urban framework. The use of resources such as OpenStreetMap (OSM) \cite{yeboah2021analysis}, and Google Street View \cite{najmi2022integrating} enriches data collection with comprehensive geographic information and panoramic street-level visuals. The crowd-sourced OSM data and the Google Street View images are beneficial for thoroughly investigating slums, as they supply a vast amount of information about urban layouts and provide detailed views of the urban environment. A progressively encouraging approach in this domain is generating synthetic training data that accurately replicates urban environments, including the intricate and diverse characteristics of slums. There are numerous benefits to using synthetic data \cite{fassnacht2018using, le2023mask}. It offers a cost-efficient alternative to gathering a large amount of real-world data. Moreover, it can be generated in significant quantities, covering a diverse range of situations and structural differences, which is essential for effectively training deep learning models. Due to its lack of specific data acquisition restrictions, this method significantly reduces the biases frequently present in conventional datasets. Synthetic datasets can also be tailored to emphasize specific attributes of slum regions, guaranteeing that models are adequately prepared to handle the wide range of variations and unique features in these settings. Combining satellite imagery with street-level data and the smart use of synthetic training data offers a complete and multi-dimensional method for mapping urban slums. This methodology overcomes the difficulties posed by the intricate characteristics of these environments and enhances the potential of deep learning models by offering them varied, abundant, and impartial data for efficient training. 

Regarding the choice of predictive models, it is important to represent the complex urban spatial structures in the feature maps. Vision Transformers (ViT), a computational model that analyses image patches and identifies distant relationships, can highlight the distinctive characteristics of urban design \cite{vit1, vit2, vit3, vit4}. Moreover, with the growing complexity of models, their reliability is an important concern. Using models that are very good at transferring across a wide range of image sources and conditions, including explainability tools, can help us understand how these models make decisions and trust their outcomes.

While our study provides comprehensive insights into the use of deep learning for slum mapping, it is important to acknowledge certain limitations. The effectiveness of the models depends on the quality and diversity of the training datasets, which may not always be available or represent all types of slum environments. Future research should focus on developing techniques for generating synthetic datasets that can accurately reflect the complex nature of slums. Additionally, integrating newer architectures and hybrid models could improve accuracy and adaptability in diverse urban contexts as deep learning evolves. There's also a need for studies focusing on technological advancement and considering the socio-economic implications, ensuring that these technologies are used responsibly and ethically.

\subsection{Limitations of Remote Sensing in Slum Mapping}

One of the fundamental challenges in the application of remote sensing for slum detection and mapping lies in the conceptual ambiguity surrounding the definition of slums \cite{habitat2003challenge}. There is a wide diversity in their appearance, both locally and globally \cite{challenges}. This leads to differences in assessments of a slum's boundaries. Such ambiguities make it difficult to train and validate algorithms, impacting their geographic, contextual, and temporal transferability. Moreover, a gap often exists between the remote sensing community and users in understanding the data requirements for different user groups and the capabilities and limitations of remote sensing. This disconnect hinders the production of maps suitable for diverse user groups, including local and global policymakers.

The limitations of remote sensing in slum detection and mapping become more pronounced when considering the difficulties in accessing basic amenities like sanitation and water \cite{schmitt2017robus}. Remote sensing provides a valuable perspective on the physical layout and extent of slums but lacks the capability to directly assess the availability and quality of essential services such as clean water and sanitation facilities. This limitation is significant, as these aspects are crucial for understanding the living conditions within slums. Remote sensing data predominantly captures topographical and physical features visible from space, but detailed information about the internal conditions of buildings, the quality of water sources, and the presence of sanitation facilities require ground-based assessments or other data sources. This gap means that while satellite imagery can identify areas likely to be slums based on physical characteristics, it cannot provide comprehensive insight into the quality of life or the health risks associated with inadequate access to clean water and proper sanitation.

The lack of high-resolution data in some regions exacerbates this challenge. Even where high-resolution imagery is available, distinguishing between different types of small-scale infrastructure, like water taps or toilets, is incredibly challenging. A combination of remote sensing data with other data sources, such as ground surveys, census data, and participatory mapping with local communities, is essential to address these issues \cite{klemmer2020population}. This multi-source approach enhances the accuracy of slum mapping and provides a more holistic view of the living conditions within these communities. By overcoming these challenges, remote sensing can significantly contribute to the understanding and improvement of living conditions in slums, aiding global and local efforts in urban planning and policy-making.

\section{Conclusion}

The rapid progress of deep learning has resulted in significant breakthroughs in the domain of slum mapping using remote sensing data. This study consolidates findings from 40 studies between 2014 and 2024, picked from a database of scholarly contributions. The cumulative results of these studies provide a comprehensive portrayal of the transformative impact of deep learning methods on our capacity to identify, examine, and comprehend slums.

The dynamic nature of CNN evolution is shown to have important consequences for slum mapping. Considering the intricate and dynamic characteristics of urban slums, it is imperative for researchers and urban planners to stay updated about the current deep-learning approaches. Nevertheless, there is a growing consensus among the community that, rather than continuously developing new architectural designs, there is considerable value in enhancing and optimizing current models to address the obstacles associated with slum mapping \cite{reviewron}. The study continues by expressing an optimistic perspective, stating that CNNs are poised to play a leading role in the future of slum mapping. This advancement is expected to enable more urban insights, building upon the current successes in this field. Within the realm of slum mapping, data serves as a representation of the complex socio-spatial structure, and the core elements for achieving success in deep learning involve ensuring the granularity, quality, and representativeness of the data. The implementation of a data-centric approach has the potential to effectively address the disparity between advanced models and the practical challenges faced in slum areas. The advancements in deep learning for slum mapping have direct implications for urban development strategies. Accurate mapping can inform infrastructure development, resource allocation, and service delivery in informal settlements. Policymakers can utilize these models to monitor urban growth, plan sustainable development projects, and prioritize interventions in the neediest areas. This meta-analysis also suggests areas where further research could be beneficial, such as developing models that are better tailored to local conditions or that can integrate data from a variety of sources. Policymakers can use these findings to guide research funding towards these areas, ensuring that future developments continue to improve the quality and utility of slum maps. 

This study culminates with a message of optimism, envisioning a future where deep learning is integral to our understanding and improvement of urban environments. For policymakers, this means supporting research initiatives that bridge the gap between high-level models and on-the-ground urban challenges. For practitioners, it means embracing these tools to foster more informed, evidence-based decision-making. In conclusion, as we harness the power of CNNs for slum mapping, we are on the verge of a new era in urban planning—one where technology enables us to see the unseen and act with greater precision for the betterment of all urban inhabitants.

\end{document}